\documentclass[10pt,twocolumn,letterpaper]{article}

\usepackage[pagenumbers]{wacv}

\usepackage{graphicx}
\usepackage{booktabs}
\usepackage{mathtools}
\usepackage{tabularx}
\usepackage{pifont} 
\usepackage{amsmath, algorithm, algorithmic}
\usepackage{multirow}
\usepackage{subcaption}
\usepackage[table]{xcolor}

\usepackage[accsupp]{axessibility}  

\usepackage{hyperref}

\usepackage{orcidlink}

\newcommand\Tstrut{\rule{0pt}{2.2ex}}         
\newcommand\Bstrut{\rule[-0.5ex]{0pt}{0pt}}   

\begin{document}

\hbadness=2000000000
\vbadness=2000000000
\hfuzz=100pt


\title{Multi-Scale Temporal Domain Alignment\\ for Federated Video Domain Adaptation} 



\author{
Lee En-Yi Hannah\textsuperscript{1}
\quad Haozhi Cao\textsuperscript{2}
\quad Yuecong Xu\textsuperscript{1}\thanks{Corresponding author.}
\\
\textsuperscript{1}{National University of Singapore} \quad
\textsuperscript{2}{The University of Tokyo}
\\
\small{\texttt{hannah.lee@u.nus.edu}} \quad \small{\texttt{h-cao@mi.t.u-tokyo.ac.jp}} \quad \small{\texttt{yc.xu@nus.edu.sg}}
}

\maketitle

\begin{abstract}
  Federated Video Domain Adaptation (FVDA) enables collaborative learning across distributed and non-IID video datasets while preserving privacy, but is under-explored due to challenges in aligning temporal information. We propose Multi-scalE Temporal domAin aLignment (METAL), a novel framework that leverages temporal information at multiple resolutions to improve cross-domain video action recognition with only model parameter transfers. METAL trains per-scale transformer encoders on source-clients, then performs independent knowledge voting at each temporal scale to generate robust pseudo-labels on the target-server. A novel $L_2$ variance penalty enforces cross-scale consistency during scale-based knowledge distillation, preventing a singular dominant scale. The late fusion aggregates features across different scales, where the fusion head is trained via knowledge distillation using confidence-weighted aggregation of scale-wise predictions, enabling the model to effectively exploit complementary temporal information for final predictions.
  Experiments on Epic-Kitchens-55 and Daily-DA demonstrate state-of-the-art performances, with gains up to 28.47\% over current FDA methods. Ablation studies prove that multi-scale distillation and scale coordination are critical for effective temporal knowledge transfer.

\end{abstract}

\section{Introduction}
\label{sec:intro}
Federated Domain Adaptation (FDA) is a privacy-preserving machine learning paradigm that seeks to transfer knowledge from the multiple source-clients to a target-server with statistically heterogeneous distributions. FDA is more robust than Federated Learning (FL) as real-world datasets are often non-IID. Without domain alignment to the target data, source models have poor out-of-domain performance and their aggregations fail to generalise to the target domain. 

While extensively explored with image data~\cite{yiFDACFederatedDomain2025, jiang2024principled, 10.1609/aaai.v38i12.29311}, FDA has yet to be studied for the more challenging video modality, which contains significantly more information, specifically the temporal information. In applications such as security surveillance, users such as homeowners or building managers who wish to benefit from an existing security system without sending their data to a third-party database can participate as a target-server. Models trained locally on other surveillance cameras can be adapted to these new users with varied scenarios (e.g., lighting and background) without access to existing local surveillance data that contains sensitive information such as human faces via applying FDA frameworks. However, current FDA methods are image-based and lack the ability to process temporal information in videos, leading to poor performance on the target domain. Hence, developing Federated Video Domain Adaptation (FVDA) would be invaluable to the creation of more robust models while preserving data privacy.

Video data presents additional challenges as domain alignment must be achieved for both spatial and temporal dimensions to develop an effective FVDA method. Specifically, to align the temporal domain, two key challenges must be addressed. (1) Temporal characteristics vary even within a video, hence temporal alignment requires considering multiple frequencies simultaneously. For instance, the action “wash” would involve rapid scrubbing motions at the wrist, but mostly stationary arms to hold the object in the sink. Recognising such action requires jointly representing motion patterns of different frequencies. Furthermore, for each frequency, the temporal characteristics are inconsistent between domains. That is, we cannot assume that the domain shift is uniform across all temporal scales. (2) Existing Video Unsupervised Domain Adaptation (VUDA) works typically approach challenge (1) by statistically aligning extracted features from the source and target domains during training~\cite{weiUnsupervisedVideoDomain2023, sacilotti_transferable-guided_2025}. However, the federated setting prohibits data access across source-clients and the target-server, which rules out direct application of VUDA techniques and further amplifies the challenges of frequency-wise feature alignment of FVDA. Moreover, most VUDA works consider a single-source-single-target scenario, whereas FVDA involves multiple source-clients with distinct data distributions. As data access across source-clients is also not permitted in FVDA, having multiple sources adds further complexity, as information from multiple varied statistical distributions must be considered during the alignment process. 

To address the challenge of temporal-domain alignment in FVDA, both the variation in temporal characteristics within the video and across domains must be addressed without cross-domain data access. Recognising that capturing temporal information at different frequencies is a prerequisite to performing temporal alignment, we take inspiration from the multi-temporal-scale works for the action recognition task~\cite{9008780}. We propose that a model with multiple pathways that analyse the video at a few selected frequencies could be used to capture the varied temporal information simultaneously. Each pathway takes in the same video at different frame rates, extracting the temporal information which is later fused to form a holistic temporal representation of the action. Then, to address the temporal domain shift on a temporal scale level in the federated setting where only model parameters are accessible, it is essential to build a more generalised model from the source-clients. That is, we hypothesise that forming a consensus amongst the source-clients will make transferring knowledge to the target-server easier. Furthermore, consensus formation at a scale level can address the individual domain shifts of each scale separately. To accomplish this, we draw inspiration from the consensus vote process in KD3A~\cite{pmlr-v139-feng21f}, a knowledge distillation-based unsupervised FDA work. The consensus vote not only creates high consensus and confidence pseudo-labels for the unlabelled target domain but also provides insight into the consensus quality across the source domains. The consensus vote process hence forms consensus knowledge, which can be used alongside the source model parameters to create a generalised source model. Per-scale generalised source models can serve as teachers to their respective scale's target model to address temporal domain shifts at a scale level. Subsequently, a fusion step can serve to aggregate temporal features from across scales to create a final prediction on the target data. This fusion step can be further guided by a fusion generalised source model. 


To this end, we propose METAL, a novel FVDA method. Instead of naively sampling videos frame-wise and applying an image-based FDA method, we consider the temporal domain shift explicitly by learning multi-temporal-scale representations of video data and handling temporal domain alignment on a temporal scale level. To learn such multi-scale representations, each source-client trains multiple models on its video data sampled at different frequencies. \textit{Per-scale knowledge distillation} is performed by creating generalised source-models, using only source model parameters, to act as a teacher for its corresponding target model. \textit{Scale coordination via $L2$ variance minimisation} prevents any singular temporal scale from dominating training, ensuring the fusion process does not overlook temporal information from any scale. Finally, a \textit{late fusion} approach is used so that distinct temporal representations are learned by each scale, preventing scales from providing overlapping information during fusion.

Our key contributions are threefold. First, we propose METAL, the first work, to the best of our knowledge, to tackle Federated Video Domain Adaptation (FVDA). Second, METAL tackles the challenge of FVDA by approaching the temporal domain shift in a multi-scale manner while leveraging only model parameters. It creates per-scale generalised source-models for per-scale knowledge distillation, utilises an $L2$ variance loss to prevent scale-dominance and late fusion with feature-based knowledge distillation to combine the distinct temporal representations for temporal domain alignment across all scales. 
Third, we demonstrate the efficacy of METAL with extensive baseline experiments. METAL outperforms current FDA methods in cross-domain action recognition datasets such as Epic-Kitchens-55~\cite{DBLP:journals/corr/abs-1804-02748} and Daily-DA~\cite{xu2023multi}, and achieves up to 28.47\% relative performance gain.

\section{Related Works}
\noindent
\textbf{Federated Domain Adaptation.}
FDA frameworks emphasise learning invariant features to increase the generalisability of the model to different domains. Previous FDA works have explored minimising adaptation losses alongside local task loss~\cite{Peng2020Federated}, sending prototypes to the server for domain alignment instead of solely for aggregation~\cite{yiFDACFederatedDomain2025, 10.1609/aaai.v38i12.29311, Peng2020Federated}, knowledge distillation methods~\cite{pmlr-v139-feng21f} and adversarial training~\cite{Peng2020Federated, 9751581} methods. Adversarial methods are computationally expensive and can align larger domain shifts but are more sensitive to unimportant features eg. backgrounds, lighting differences~\cite{shaoAdversarialAutoencoderUnsupervised2019}. Knowledge distillation methods are significantly more lightweight but may suffer a lack of granularity. Aggregation of either prototypes or gradients requires attention to data privacy leaks and require good feature extractors. 
To imitate realistic settings, more FDA works have begun exploring increasingly challenging limitations. Unsupervised FDA, where target data is unlabelled, is a popular setting. Techniques such as clustering~\cite{10943413, 10030428} or creating pseudo-labels~\cite{pmlr-v139-feng21f} using the source-trained models are combined with FDA techniques to address unsupervised FDA. 
Knowledge Distillation based Decentralized Domain Adaptation (KD3A)~\cite{pmlr-v139-feng21f} is an unsupervised image FDA framework that creates consensus-based pseudo-labels for the target domain and uses knowledge distillation to teach the student target model to learn this consensus knowledge. To form the consensus knowledge, all source models participate in the knowledge vote, where they perform inference on the target data and confident and high consensus predictions area used to form the pseudo-labels. Models that contribute poorly to the consensus are given less weight in future aggregation rounds.

\noindent
\textbf{Video Unsupervised Domain Adaptation.}
Video DA aims to address the domain shift problem in the video modality, increasing the generalisability of models. Compared to image DA works, it presents the additional challenge of aligning the temporal domains of the source and target~\cite{weiUnsupervisedVideoDomain2023}. Earlier works such as TA3N~\cite{9008391} emphasised the importance of temporal feature alignment and also selecting more temporal features that contribute more to domain shift using attention mechanisms. This improved performance over naïve methods that sampled video frames and applied image DA methods directly, which ignored temporal domain shift. Domain alignment for spatiotemporal features has been achieved through adversarial methods~\cite{9008391}, contrastive learning~\cite{da_Costa_2022_WACV} and attention~\cite{10.1007/978-3-030-58610-2_40, sacilotti_transferable-guided_2025}. Subsequently, other methods of handling spatio-temporal features have emerged. TranSVAE~\cite{weiUnsupervisedVideoDomain2023} tackles video DA from a disentanglement view. By extracting the static and dynamic latent factors from videos and designing specific objectives for each of them, TranSVAE achieves better domain alignment in both spatial and temporal domains. \cite{10.1007/978-3-030-58610-2_40} proposed an attention mechanism to place greater emphasis on clips with discriminative information instead of equally aligning all clips. In sum, some notable innovations in video DA include techniques to align temporal domains, the creation of specific objectives for temporal features and methods to remove focus from the uninformative details in videos. Most video DA works have assumed the closed-set video unsupervised DA setting defined in~\cite{10.1145/3679010}. However, the federated setting violates most of these assumptions and prevents direct application of video DA methods to FDA.

\noindent
\textbf{Temporal Representation for Videos.}
Multi-Temporal-Scale networks were first proposed in Feichtenhofer et al.’s SlowFast~\cite{9008780}. SlowFast had two pathways with different temporal resolutions, allowing the slower pathway to focus on capturing semantic information while the faster pathway captured rapid motion. This trains a model with a dual understanding of the spatiotemporal information in videos. Variations of multi-scale temporal modelling have since been developed. Pathformer uses adaptive temporal scales to suit the temporal characteristics of the dataset~\cite{chen2024pathformer} instead of fixed temporal scales, MSTCN performs scale fusion using attention mechanisms that reduce redundancies across scales~\cite{XU2024110288} and MTT leverages lateral cross attention mechanisms across scales to better align semantic information~\cite{9681250}. 

\section{Proposed Method}
\begin{figure}[t]
    \centering
    \includegraphics[width=1.\linewidth]{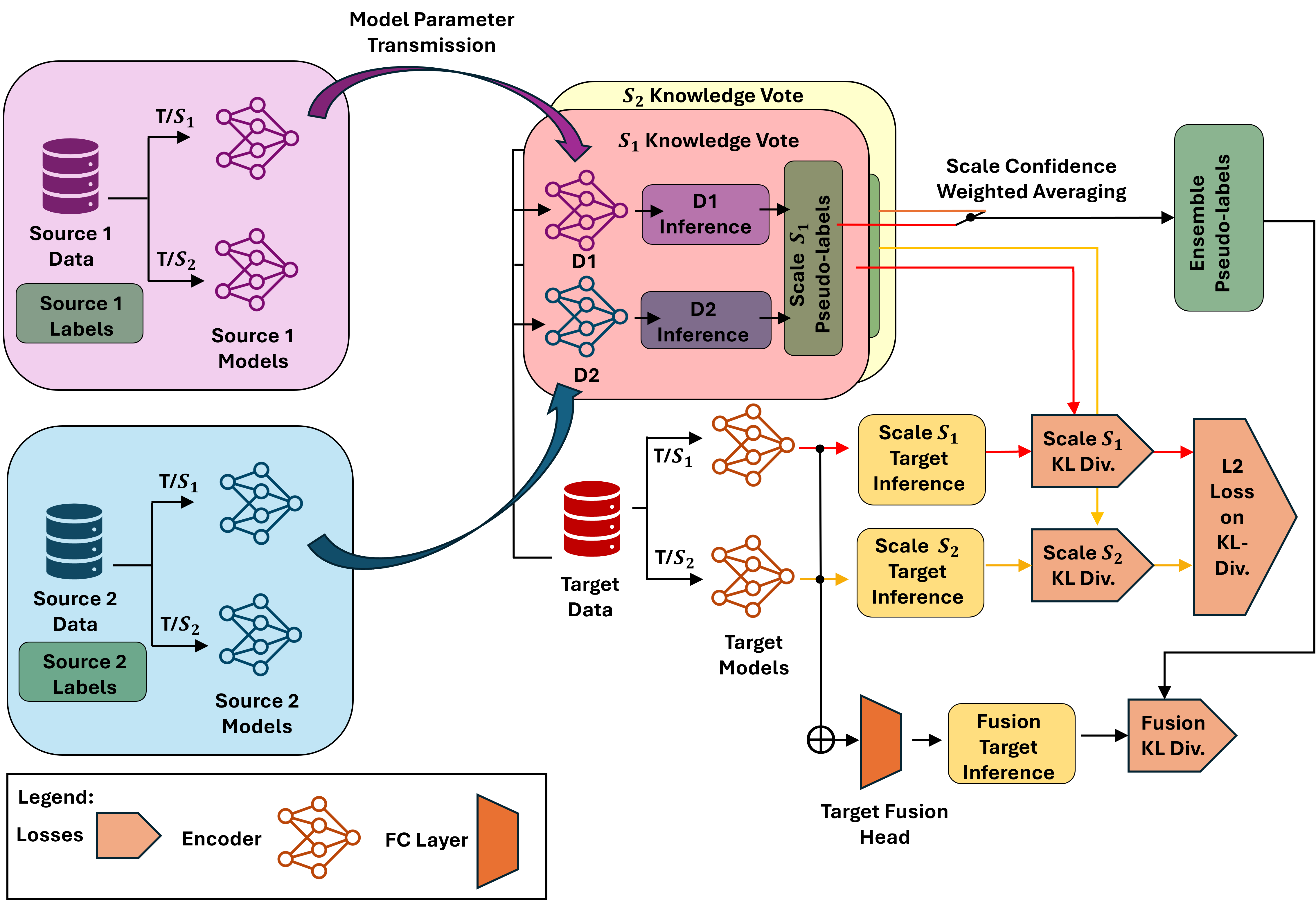}
    \caption{\textbf{METAL Architecture.} Source-clients train per-scale models independently before uploading to the target-server. For each temporal scale, a knowledge vote process aggregates source predictions into high-confidence pseudo-labels, forming a generalised teacher model. Target students minimize per-scale KL divergence with these pseudo-labels, while an $L2$ variance penalty enforces cross-scale consistency. The target fusion head is trained on concatenated multi-scale target features, distilled using confidence-weighted ensemble pseudo-labels as the teacher signal. After each communication round, federated averaging with consensus-based domain weights updates all models. Only 2 scales are shown for clarity.}
    \label{fig:architectureDiag}
\end{figure}
For Federated Video Domain Adaptation (FVDA), we consider the Multi-Source-Single-Target scenario with $K$ source domains $\{D^k_S\}_{k=1}^K$ each contained within a source-client and target domain $D_T$ on a target-server. Each source domain contains $N_k$ labelled videos and the target domain contains $N_T$ unlabelled videos (i.e., $D^k_S\coloneq \{X_i^k,y_i^k\}^{N_k}_{i=1}$, $D_T\coloneq\{X_i^T, y^T_i\}^{N_T}_{i=1}$ where $y^T_i$ is unobserved). Each domain is drawn from its corresponding distribution $P_S^k (x,y)$ and $P_T(x,y)$ with domain shift, $P_S^k\neq P_T$. We assume that both the labelled source-client videos and the unlabelled target-server videos share the same $C$ classes. For the federated setting, communication between source-clients and target-server is limited to model parameters, i.e., $\{D^k_S\}_{k=1}^K$ is inaccessible when adapting the source models to $D_{T}$.

\subsection{Multi-Scale Temporal Feature Representation}
We represent an action segment $X$ as a temporal sequence of frame-features $\mathbf{F} = \{f_1, f_2, \ldots, f_T\} \in \mathbb{R}^{T \times d}$, where $T$ is the number of temporal segments and $d$ is the feature dimension. Each frame-feature is constructed from a clip centred at the frame of interest in order to preserve temporal information at the feature extraction stage. To capture temporal dynamics at multiple scales, we construct $V$ temporal views of each action segment, aiming to capture a range of motion patterns from fine-grained to coarse-level action patterns. 
For each scale $s \in \mathcal{S}, \mathcal{S}=\{S_i\}^{V}_{i=1}$, we uniformly sample $T/s$ frames from the original sequence to construct the multi-scale representation $\{\mathbf{F}^{(s)}\}_{S}$. This multi-scale temporal sampling enables the model to learn scale-invariant features that are robust to temporal variations across domains.

\subsection{Source Model Training}
\subsubsection{Per-Scale Encoder Architecture.}
\label{ref:transformer_encoder}

For each temporal scale in $S$, we employ a transformer-based encoder $\mathcal{E}^{(s)}: \mathbb{R}^{T/s \times d} \rightarrow \mathbb{R}^{D}$ to extract domain-specific representations. The encoder processes the input features through $\mathcal{E}^{(s)}(\mathbf{F}^{(s)})=\text{LayerNorm}(\text{CLS}(\mathcal{H}^{(s)}(\mathbf{F}^{(s)})))$, where $\mathcal{H}^{(s)}$ is multi-layer transformer encoder and $D$ is the hidden dimension. Each scale-specific encoder is paired with a classifier $\mathcal{C}^{(s)}: \mathbb{R}^D \rightarrow \mathbb{R}^C$ that predicts action labels.

\subsubsection{Source Training Objective.}
Each source client $k$ independently trains its scale-specific models using the standard cross-entropy loss:
\begin{equation}
\mathcal{L}_{\text{source}}^{(s,k)} = -\frac{1}{N_k}\sum_{i=1}^{N_k} \sum_{c=1}^{C} y_{i,c}^k \log p_{i,c}^{(s,k)}, 
\end{equation}
where $y_{i,c}^k$ is the one-hot encoded label and $p_{i,c}^{(s,k)}$ is the predicted probability for class $c$. Each scale is trained independently $\mathcal{L}_{\text{source}}^k = \sum_{S} \mathcal{L}_{\text{source}}^{(s,k)}$. After training, each source client uploads the parameters of its three trained models $\{\theta_{\mathcal{E}^{(s)}_k}, \theta_{\mathcal{C}^{(s)}_k}\}_{S}$ to the target server.

\subsection{Per-Scale Knowledge Distillation}

\subsubsection{Per-Scale Generalised Source Model Creation.}
\noindent
To facilitate knowledge transfer to the target domain, we aim to create a generalised source model for each scale that will act as a teacher to the target model. Naively averaging source model parameters is not feasible due to the domain shift; instead, we adapt K3DA's~\cite{pmlr-v139-feng21f} knowledge vote and perform \textit{separate knowledge voting for each temporal scale}. This is crucial as different scales may have different levels of consensus across source domains. The knowledge vote outputs high-confidence and high-consensus pseudo-labels for each scale. This meaningfully combines the inferences from the individual source domains and allows them to behave as a single generalised source model which is used to teach its corresponding scale target model as seen in Fig.~\ref{fig:architectureDiag}. 
For a target sample $X_i^T$ at scale $s$, each source domain $k$ produces a prediction $p_k^{(s)}(X_i^T) = \text{softmax}(\mathcal{C}_k^{(s)}(\mathcal{E}_k^{(s)}(\mathbf{F}_i^{(s,T)}))).$ We then collect predictions from all $K$ source domains to form a knowledge matrix $\mathbf{P}^{(s)} \in \mathbb{R}^{B \times K \times C}$ for batch size $B$. We then perform knowledge voting to identify confident predictions. For each sample, we compute the consensus by majority voting $\hat{y}_i^{(s)} = \text{argmax}_c \sum_{k=1}^{K} \mathbb{I}[\text{argmax}(p_k^{(s)}(X_i^T)) = c]$, where $\mathbb{I}[\cdot]$ is the indicator function. The consensus confidence is measured by:
\begin{equation}
\text{conf}_i^{(s)} = \frac{\max_c \sum_{k=1}^{K} \mathbb{I}[\text{argmax}(p_k^{(s)}(X_i^T)) = c]}{K}.
\label{eq: per_scale_consensus_confidence}
\end{equation}

Samples with confidence above a threshold $\tau$ are considered reliable for pseudo-labeling. The consensus knowledge for scale $s$ is:
\begin{equation}
q^{(s)}(X_i^T) = \frac{1}{|\mathcal{A}_i^{(s)}|} \sum_{k \in \mathcal{A}_i^{(s)}} p_k^{(s)}(X_i^T),
\label{eq: per_scale_consensus_knowledge}
\end{equation}
where $\mathcal{A}_i^{(s)} = \{k : \text{argmax}(p_k^{(s)}(X_i^T)) = \hat{y}_i^{(s)}\}$ is the set of agreeing source domains. The confidence gate $\tau$ is linearly annealed from $\tau_{\text{begin}}$ to $\tau_{\text{end}}$ during training as too strict a confidence a gate in early epochs will eliminate too many domains from the knowledge vote, resulting in no consensus being formed.

To further enhance model robustness to the domain shifts, we apply Mixup~\cite{zhang2018mixup} augmentation with $\lambda \sim \text{Beta}(2, 2)$ to both inputs and pseudo-labels to obtain $\tilde{X}_i^T$ and $\tilde{q}^{(s)}(X_i^T)$, respectively. This creates soft interpolations between samples, improving generalisation.

\subsubsection{Per-Scale Knowledge Distillation Loss.}
\noindent
For each scale $s$, we train a student model $\{\mathcal{E}_T^{(s)}, \mathcal{C}_T^{(s)}\}$ to minimize the KL divergence between its predictions and the consensus pseudo-label:
\begin{equation}
\mathcal{L}_{\text{KL}}^{(s)} = \frac{1}{B}\sum_{i=1}^{B} w_i^{(s)} \cdot D_{\text{KL}}(p_T^{(s)}(\tilde{X}_i^T) \,\|\, \tilde{q}^{(s)}(X_i^T))
\label{eq: per_scale_kl_div}
\end{equation}
where $w_i^{(s)} = \mathbb{I}[\text{conf}_i^{(s)} > \tau]$ is a binary weight indicating reliable samples, and $D_\mathrm{KL}(\cdot \, \| \, \cdot)$ denotes the KL divergences. 

\subsubsection{Scale Coordination via $L2$ Variance Minimisation.}
An $L2$ variance penalty that enforces consistent learning across temporal scales is introduced. We observe that different scales may learn at different rates, leading to imbalanced representations which cause our model to overlook key temporal features present in other frequencies. To address this, the variance of per-scale KL divergence losses is minimised with:
\begin{equation}
\mathcal{L}_{\text{var}} = \sqrt{\frac{1}{3}\sum_{S} \left(\mathcal{L}_{\text{KL}}^{(s)} - \bar{\mathcal{L}}_{\text{KL}}\right)^2}, 
\label{eq: L2_loss}
\end{equation}
where $\bar{\mathcal{L}}_{\text{KL}} = \frac{1}{3}\sum_{S} \mathcal{L}_{\text{KL}}^{(s)}$ is the mean KL loss across scales. The square root ensures the penalty is in the same scale as the KL losses. This regularisation prevents any single scale from dominating the learning process and encourages balanced knowledge transfer across all temporal resolutions.

The total loss for target adaptation combines per-scale distillation with scale coordination:
\begin{equation}
\mathcal{L}_{\text{adapt}} = \sum_{S} \mathcal{L}_{\text{KL}}^{(s)} + \beta \mathcal{L}_{\text{var}},
\label{eq: adapt_loss}
\end{equation}
where $\beta$ is a hyper-parameter controlling the strength of scale coordination. Critically, this loss is computed with gradients flowing through all scale models simultaneously. During backpropagation, the variance term $\mathcal{L}_{\text{var}}$ connects all scales, ensuring that the gradient updates consider inter-scale consistency. 

\subsection{Scale Fusion}
\noindent
After training the scale-specific student models, we perform late fusion to combine information from all temporal scales. A late fusion approach allows the distinct temporal representations of each scale to be independently learned in contrast to introducing lateral connections earlier in training which may cause features output by each scale to provide overlapping information to the fusion head. We train a separate fusion classifier $\mathcal{C}_{\text{fuse}}$ that takes as input the concatenated features from all scales $\mathbf{z}_{\text{fuse}} = [\mathcal{E}_T^{(S_1)}(\mathbf{F}^{(S_1,T)});...; \mathcal{E}_T^{(S_3)}(\mathbf{F}^{(S_3,T)})] \in \mathbb{R}^{VD}$ via feature-based knowledge distillation. 
The fusion classifier is trained using an ensemble pseudo-label that combines the consensus from all scales, weighted by their respective confidences:
\begin{equation}
\tilde{q}_{\text{fuse}}(X_i^T) = \frac{\sum_{S} w_i^{(s)} \tilde{q}^{(s)}(X_i^T)}{\sum_{S} w_i^{(s)} + \epsilon}, 
\label{eq: ensemble_pseudolabel}
\end{equation}
where $\epsilon = 10^{-8}$ prevents division by zero. The fusion loss is:
\begin{equation}
\mathcal{L}_{\text{fusion}} = \frac{1}{B}\sum_{i=1}^{B} \max_s(w_i^{(s)}) \cdot D_{\text{KL}}(p_{\text{fuse}}(\tilde{X}_i^T) \,\|\, \tilde{q}_{\text{fuse}}(X_i^T)), 
\label{eq: fusion_loss}
\end{equation}
where $\max_s(w_i^{(s)})$ indicates whether the sample has reliable consensus from \textit{any} scale and $ p_{\text{fuse}}(\tilde{X}_i^T) = \mathcal{C}_{\text{fuse}}(\mathbf{z}_{\text{fuse}} )$. This union strategy allows the fusion classifier to leverage confident predictions from any temporal scale. 

To adaptively weigh the contribution of each source domain, we employ consensus focus~\cite{pmlr-v139-feng21f}. During training, we track how often each source domain agrees with the final consensus prediction. The consensus focus score for source domain $k$ is computed as:
\begin{equation}
\phi_k = \frac{1}{B \cdot E} \sum_{e=1}^{E} \sum_{i=1}^{B} \mathbb{I}[\text{argmax}(p_k(X_i^T)) = \text{argmax}(\bar{q}(X_i^T))],
\end{equation}
where $E$ is the number of batches processed and $\bar{q}(X_i^T)$ is the consensus prediction (averaged across scales for domain weighting). These scores are normalised to obtain domain weights $w_k = \frac{\phi_k}{\sum_{j=1}^{K}\phi_j}$. 

After each communication round, we perform federated averaging to update all models. For each scale $s$ independently, we aggregate the encoder and classifier parameters:
\begin{equation}
\theta_{\mathcal{E}^{(s)}}^{t+1} = \sum_{k=0}^{K} w_k^t \theta_{\mathcal{E}_k^{(s)}}^t, \quad \theta_{\mathcal{C}^{(s)}}^{t+1} = \sum_{k=0}^{K} w_k^t \theta_{\mathcal{C}_k^{(s)}}^t, 
\end{equation}
where $t$ indexes the communication round, $w_0^t$ is the target domain weight (based on the proportion of confident predictions), and $\{w_k^t\}_{k=1}^K$ are the source domain weights from consensus focus. The fusion classifier is aggregated similarly:
$\theta_{\mathcal{C}_{\text{fuse}}}^{t+1} = \sum_{k=0}^{K} w_k^t \theta_{\mathcal{C}_{\text{fuse},k}}^t$. This ensures that more informative domains contribute more to the target model while maintaining privacy through parameter-only communication. The complete training procedure is summarised in Algorithm~\ref{alg:METAL}.

\begin{algorithm}[!ht]
\caption{METAL for Federated Video Domain Adaptation}
\label{alg:METAL}
\begin{algorithmic}[1]
\REQUIRE Source domains $\{D_S^k\}_{k=1}^K$, target domain $D_T$, scales $\mathcal{S}=\{S_i\}^{V}_{i=1}$, confidence threshold schedule $\{\tau_t\}$, $L2$ weight $\beta$
\ENSURE Adapted models $\{\mathcal{E}_T^{(s)}, \mathcal{C}_T^{(s)}\}_{s \in \mathcal{S}}$ and fusion classifier $\mathcal{C}_{\text{fuse}}$

\STATE \textbf{Phase 1: Source Training (on clients $k=1,\ldots,K$ in parallel)}
\STATE Train per-scale models $\{\mathcal{E}_k^{(s)}, \mathcal{C}_k^{(s)}\}_{s \in \mathcal{S}}$ and fusion $\mathcal{C}_{\text{fuse},k}$ on $D_S^k$.Upload all parameters to server

\STATE \textbf{Phase 2: Multi-Scale Target Adaptation (on server)}
\STATE Initialize: Aggregate source models, set $\phi_k \leftarrow 0$, $\tau_t$ schedule

\FOR{each communication round $t$}
    \FOR{each target batch $\{X_i^T\}$ from $D_T$}
        \FOR{each scale $s \in \mathcal{S}$}
            \STATE Create Per-Scale generalised source models \hfill \textit{(Eq. \ref{eq: per_scale_consensus_confidence}-\ref{eq: per_scale_consensus_knowledge})}
            \STATE Calculate: $\mathcal{L}_{\text{KL}}^{(s)}$ \hfill \textit{(Eq. \ref{eq: per_scale_kl_div})}
        \ENDFOR
        \STATE Add $L2$ Variance Loss: $\mathcal{L}_{\text{adapt}} = \sum_{s \in \mathcal{S}} \mathcal{L}_{\text{KL}}^{(s)} + \lambda \mathcal{L}_{\text{var}}$ \hfill \textit{(Eq. \ref{eq: L2_loss}-\ref{eq: adapt_loss})}
        \STATE \textbf{Joint backward:} $\theta_T^{(s)} \leftarrow \theta_T^{(s)} - \alpha \nabla_{\theta_T^{(s)}} \mathcal{L}_{\text{adapt}}$ for all $s$ \textit{(gradients connect scales)}
        \STATE Ensemble pseudo-label: $\tilde{q}_{\text{fuse}}$ \hfill \textit{(Eq.~\ref{eq: ensemble_pseudolabel})}
        \STATE Train: $\mathcal{L}_{\text{fusion}} = \frac{1}{B}\sum_i \max_s(w_i^{(s)}) D_{\text{KL}}(p_{\text{fuse}}(\tilde{X}_i^T) \| \tilde{q}_{\text{fuse}})$ \hfill \textit{(Eq.~\ref{eq: fusion_loss})}
        
        \STATE Update consensus focus $\phi_k$ and compute domain weights $w_k$
    \ENDFOR
    
    \STATE \textbf{Federated aggregation:} $\theta^{t+1} = \sum_{k=0}^K w_k^t \theta_k^t$ for all scales and fusion
\ENDFOR

\RETURN $\{\mathcal{E}_T^{(s)}, \mathcal{C}_T^{(s)}\}_{s \in \mathcal{S}}, \mathcal{C}_{\text{fuse}}$
\end{algorithmic}
\end{algorithm}

\section{Experiments}
In this section, we evaluate METAL across two widely used cross-domain action recognition benchmarks: Epic-Kitchens-55~\cite{DBLP:journals/corr/abs-1804-02748} and Daily-DA~\cite{xu2023multi}, under the federated domain adaptation setting. Ablation studies, hyper-parameter sensitivity studies and empirical analysis of METAL are also presented to validate our proposed architecture. 

\begin{table}[ht]
\centering
\caption{Sensitivity analysis of the number of temporal segments ($T$) on Epic-Kitchens-55. Performance scales significantly with higher temporal resolution.}
\label{tab:segments_sensitivity}
\small
\resizebox{\linewidth}{!}{
\begin{tabularx}{\linewidth}{p{2.5cm}XXXXc}
\toprule
\textbf{Num. Segments ($T$)} & \textbf{Scale 1} & \textbf{Scale 4} & \textbf{Scale 16} & \textbf{Target Acc.} \\ \midrule
16  & 0.3101 & 0.3162 & 0.3029 & 0.3223 \\
64  & 0.4240 & 0.4254 & 0.4230 & 0.4158 \\
\rowcolor[gray]{0.9} 
\textbf{256 (Ours)} & \textbf{0.5216} & \textbf{0.4990} & \textbf{0.5082} & 0.5298 \\
512 & 0.5226 & 0.5421 & 0.5164 & \textbf{0.5380} \\ \bottomrule
\end{tabularx}
}
\end{table}

\subsection{Experimental Settings}
\subsubsection{Datasets.}
The Epic-Kitchens-55\cite{DBLP:journals/corr/abs-1804-02748} dataset is a large egocentric dataset for the HAR task that contains recordings of daily kitchen activities collected from 32 individuals in 4 cities for a total of 55 hours. Video lengths, action sequences and the distribution of action classes across videos vary greatly, adding challenge to the dataset and making it a widely used benchmark dataset for the video HAR task. The dataset was not originally intended for domain adaptation tasks but was conveniently recorded by individuals with differing kitchens and mannerisms, creating a domain shift between the data from each participant. Following~\cite{weiUnsupervisedVideoDomain2023, 9022205}, 3 large and distinct kitchens were selected to form 3 domains with 8 shared action classes. Of these domains, 2 will serve as source domains while the third will be the target domain, resulting in three different cross-domain tasks.

Meanwhile, Daily-DA~\cite{xu2023multi} is a challenging video domain adaptation dataset that combines originally distinct video datasets. The subset we will be using comprises of 8 shared classes across ARID~\cite{xu2021arid}, HMDB51~\cite{kuehne2011hmdb} and Moments-In-Time (MIT)~\cite{monfort2019moments}. HMDB51 and MIT have widely been used in action recognition benchmarks and ARID is a dark dataset that provides additional challenge to Daily-DA with its adverse illumination conditions. This subset of Daily-DA shares 8 classes which will be used in our experiments, each dataset serves as distinct domain with 4507 videos for ARID, 1527 for HMDB51, and 4400 for MIT. A train-test split of 70-30 implemented for ARID and HMDB51, while MIT's train-test split was used. For both Epic-Kitchens and Daily-DA, we report the top-1 accuracy obtained on each setting.

\subsubsection{Detailed Implementation.}
\label{sec: detailed_implementation}
Frame-features were extracted from each video using a frozen I3D~\cite{8099985} feature extractor pre-trained on the ImageNet dataset~\cite{5206848}. Clips consist of 16 frames i.e., for each frame, the previous 7 frames and the next 8 frames are combined with the frame to form a clip. This clip is then fed to the feature extractor. Zero-padding is used for the start and end of the videos. Pre-extracted features created by~\cite{weiUnsupervisedVideoDomain2023} were used for the Epic-Kitchens-55 experiments whereas the Daily-DA features were extracted by us. Frame-features are of dimension $d=2048$. The per-scale transformer encoder utilised in all experiments is a 4-layer transformer encoder with 8 attention heads, hidden dimension $D=512$, and feed-forward dimension of 1024. We select the number of temporal segments $T =256$ to be fed into our model after experimentation as shown in Table~\ref{tab:segments_sensitivity}. Selecting a suitable $T$-value is vital as our method relies on learning temporal representations at different temporal frequencies. Small $T$ values result in the loss of critical longer-range temporal information. This is evidenced by the significant performance improvement from $T=64$ to $T=256$. As performance improvement with further increase to $T=512$ is minimal, we choose $T=256$ for our experiments to reduce computational costs. Scales $s \in \{S_1=1, S_2=4, S_3=16\}$ were also chosen after experimentation. 

\subsection{Overall Results and Comparisons}
\begin{table}[!t]
\centering
\caption{Comparison of METAL against current FDA methods on Epic-Kitchens-55.}
\label{tab:FDA_FVDA_comparison_EpicKitchens}
\footnotesize 
\setlength{\tabcolsep}{5pt} 
\resizebox{\linewidth}{!}{
\begin{tabular}{l|c|ccccc}
\hline \hline
\textbf{Source} & \textbf{Target} & \textbf{KD3A} & \textbf{FDAC} & \textbf{FedGP} & \textbf{Co-MDA} & \textbf{METAL} \Tstrut\Bstrut\\
\hline
P08, P01 & P22 & 0.4333 & 0.4466 & 0.4698 & 0.3921 & \textbf{0.5298} \Tstrut\\
P22, P01 & P08 & 0.3517 & 0.3425 & 0.4351 & 0.3597 & \textbf{0.4506} \\
P22, P08 & P01 & 0.4560 & 0.4667 & 0.4549 & 0.3963 & \textbf{0.4947} \Bstrut\\
\hline
\rowcolor[gray]{0.95} \multicolumn{2}{l|}{\textbf{Avg. Target Acc.}} & 0.4137 & 0.4186 & 0.4533 & 0.3827 & \textbf{0.4917} \\
\rowcolor[gray]{0.95} \multicolumn{2}{l|}{\textbf{Avg. METAL Perf. Gain}} & 0.1886 & 0.1746 & 0.0848 & 0.2847 & -- \\
\hline \hline
\end{tabular}
}
\end{table}

\begin{table}[!t]
\centering
\caption{Comparison of METAL against current FDA methods on Daily-DA.}
\label{tab:FDA_FVDA_comparison_DailyDA}
\footnotesize 
\setlength{\tabcolsep}{5pt} 
\resizebox{\linewidth}{!}{
\begin{tabular}{l|c|ccccc}
\hline \hline
\textbf{Source} & \textbf{Target} & \textbf{KD3A} & \textbf{FDAC} & \textbf{Co-MDA} & \textbf{METAL} \Tstrut\Bstrut\\
\hline
HMDB51, MIT & ARID & 0.2490 & 0.1979 & 0.2284 & \textbf{0.2999} \Tstrut\\
ARID, MIT & HMDB51 & 0.5930 & 0.5558 & 0.5814 & \textbf{0.5967} \\
ARID, HMDB51 & MIT & 0.3250 & 0.3229 & 0.3912 & \textbf{0.3452} \Bstrut\\
\hline
\rowcolor[gray]{0.95} \multicolumn{2}{l|}{\textbf{Avg. Target Acc.}} & 0.3890 & 0.3589 & 0.4003 & \textbf{0.4139} \\
\rowcolor[gray]{0.95} \multicolumn{2}{l|}{\textbf{Avg. METAL Perf. Gain}} & 0.0641 & 0.1535 & 0.0339 & -- \\
\hline \hline
\end{tabular}
}
\vspace{-5pt}
\end{table}

We compare METAL with state-of-the-art (SOTA) FDA approaches as shown in Table~\ref{tab:FDA_FVDA_comparison_EpicKitchens} and~\ref{tab:FDA_FVDA_comparison_DailyDA}. We use the same transformer encoder described in Section~\ref{ref:transformer_encoder} and with identical settings for the encoder across all experiments. The top-1 target accuracies are reported for each approach. The current FDA approaches include: KD3A~\cite{pmlr-v139-feng21f}, FDAC~\cite{yiFDACFederatedDomain2025}, FedGP~\cite{jiang2024principled} and Co-MDA~\cite{10128163}. Of the 4 FDA methods, FedGP considers the supervised FDA case while the rest are unsupervised FDA methods. 
KD3A~\cite{pmlr-v139-feng21f} is a knowledge distillation-based method that aims to reduce negative transfer and minimise communication costs. It utilises the knowledge vote and consensus formation process to determine which domains are contributing poorly and assigns a smaller weight to them in aggregation rounds. It performs well even with reduced communication rounds in contrast to adversarial methods. FDAC~\cite{yiFDACFederatedDomain2025} is a dual-contrastive method that aims to extract domain-invariant features and also identify class-discriminative information. The simultaneous contrastive learning goals ensure that learning is not skewed towards any source domain, which in turn improves generalisability. Co-MDA~\cite{10128163} introduces a multi-domain attention mechanism that weights source domains with greater inter-class discriminative predictions more strongly and also trains two networks that perform co-teaching. This reduces the impact of negative transfer and also any confirmation bias that a single network might learn. Finally, FedGP~\cite{jiang2024principled} is a gradient projection aggregation method that extracts useful components of source gradients, guided by the target gradient, and aggregates them with an auto-weighting scheme while filtering out negative source gradients. It considers data-scarce scenarios but assumes that the target labels are available. While these 4 works introduce highly effective mechanisms to tackle domain alignment, they fail to consider the temporal domain shift which must be addressed explicitly for video data. 

METAL outperforms all current FDA methods on Epic-Kitchens-55, achieving an average relative performance gain of 18.87\% as seen in Table~\ref{tab:FDA_FVDA_comparison_EpicKitchens}, and up to 28.47\% when compared with Co-MDA. This demonstrates the effectiveness of learning multi-scale representations of video data and handling temporal domain alignment on a scale level. 

On the more challenging Daily-DA dataset, we achieve an average performance gain of 8.38\%, outperforming all current FDA baselines except Co-MDA with MIT as the target domain. As previously explored in Section~\ref{sec: detailed_implementation}, the number of segments chosen has significant impact on model performance. Compared to Epic-Kitchens, the MIT dataset has fixed video lengths of 3s~\cite{monfort2019moments} which yield about 90 frame-features, resulting in many zero-padded frame-features which dampened METAL's performance on the MIT split. The number of segments was not lowered for the Daily-DA experiments to present an apple-to-apple comparison with the Epic-Kitchens-55 dataset and also to capture the longer range temporal information in the HMDB51 and ARID datasets, where METAL outperformed all current FDA methods.

\subsection{Ablation and Hyper-Parameter Sensitivity Study}

\begin{table}[!t]
\tiny
\center
\caption{Comprehensive ablation studies for METAL on Epic-Kitchens-55, averaged across all three tasks.}
\subfloat[Individual Scale Contribution to Fusion \label{tab:scale_ablation}]{
\resizebox{.9\linewidth}{!}{
\begin{tabular}[b]{l|ccc|c}
    \hline 
    \hline
    \textbf{Fusion} & \textbf{S1} & \textbf{S4} & \textbf{S16} & \textbf{Acc.} \Tstrut\Bstrut\\
    \hline
    1+4  & .4745 & .4734 & & .4764 \Tstrut\\
    1+16 & .4788 & & .4846 & .4780 \\
    4+16 & & .4772 & .4599 & .4677 \\
    \hline
    \rowcolor[gray]{0.95} \textbf{All} & \textbf{.4746} & \textbf{.4723} & \textbf{.4760} & \textbf{.4917} \\
    \hline 
    \hline
\end{tabular}
}
}
\vspace{6pt}
\
\subfloat[Fusion Strategy\label{tab:fusion_strategy}]{
\resizebox{.9\linewidth}{!}{
\begin{tabular}[b]{l|ccc|c}
    \hline
    \hline
    \textbf{Strategy} & \textbf{S1} & \textbf{S4} & \textbf{S16} & \textbf{Fusion} \Tstrut\Bstrut\\
    \hline
    Early & .4574 & .4711 & .4918 & .4667 \Tstrut\\
    \hline
    \rowcolor[gray]{0.95} \textbf{Late} & \textbf{.4746} & \textbf{.4723} & \textbf{.4760} & \textbf{.4917} \\
    \hline
    \hline
\end{tabular}
}
}
\vspace{6pt}
\
\subfloat[$L2$ Loss Inclusion\label{tab:l2_ablataion}]{
\resizebox{.5\linewidth}{!}{
\begin{tabular}{l|c}
    \hline
    \hline
    \textbf{Configuration} & \textbf{Acc.} \Tstrut\Bstrut\\
    \hline
    Without $L_2$ Loss & .4797 \Tstrut\\
    With $L_2$ Loss    & \textbf{.4917} \\
    \hline
    \hline
\end{tabular}
}
}
\label{tab:all_ablations}
\vspace{-10pt}
\end{table}

To further validate the architecture of METAL, we conduct our ablation studies on key components of our model. First, we investigate the contributions of each scale chosen using a Leave-One-Out approach. Second, we show the superiority of the late fusion approach which allows distinct temporal representations to be learned by each scale for better fusion performance. Finally, we prove the need for scale coordination with the $L2$ loss to prevent dominance of any scale. All ablation studies are performed over Epic-Kitchens-55, and averaged across the three tasks.

\begin{table*}[h]
\centering
\caption{Hyperparameter sensitivity study for METAL on Epic-Kitchens-55. Scale accuracies $S_1, S_2, S_3$ correspond to the three temporal resolutions selected for each configuration.}
\label{tab:hyper_sensitivity}
\small
\resizebox{.88\textwidth}{!}{
\begin{tabularx}{.95\textwidth}{p{4.8cm}p{4.8cm}cccc}
\toprule
\textbf{Parameter} & \textbf{Value} & \textbf{$S_1$} & \textbf{$S_2$} & \textbf{$S_3$} & \textbf{Target Acc.} \\ \midrule
\textbf{Scale Spacing} & [1, 2, 4] & 0.5092 & 0.5051 & 0.5237 & 0.5257 \\
\rowcolor[gray]{0.9} & \textbf{[1, 4, 16] (Ours)} & \textbf{0.5216} & \textbf{0.4990} & \textbf{0.5082} & \textbf{0.5298} \\
& [1, 8, 64] & 0.5246 & 0.5185 & 0.4004 & 0.5267 \\ \midrule
\textbf{$L2$ Loss Weight} & 0.01 & 0.5257 & 0.5103 & 0.4867 & 0.5267 \\
& 0.05 & 0.5277 & 0.5072 & 0.4846 & 0.5267 \\
\rowcolor[gray]{0.9} & \textbf{0.10 (Ours)} & \textbf{0.5216} & \textbf{0.4990} & \textbf{0.5082} & \textbf{0.5298} \\
& 0.20 & 0.5154 & 0.5062 & 0.4856 & 0.5267 \\
& 0.50 & 0.5175 & 0.5041 & 0.5010 & 0.5164 \\ \bottomrule
\end{tabularx}
}
\end{table*}

\subsubsection{Individual scale contribution.} Each scale learns distinct temporal representations of the action and contributes unique information to the fusion head. Thus, ablation of any scale should result in performance loss as the model loses a frequency range of temporal information. Indeed, using all 3 scales achieved a higher fusion target accuracy than any subset of the scales, shown in Table~\ref{tab:scale_ablation}. This shows that the individual scales have succeeded in learning distinct temporal representations which allows the model to fuse the representations for a greater target accuracy than any individual scale achieves.

\subsubsection{Fusion Method.}
Late fusion allows scales to specialise and learn temporal representations at a designated frequency. In contrast, early fusion encourages contextual information sharing, temporal representations learned would blend across scales. This would work against the per-scale knowledge distillation that addresses the individual temporal domain shifts of each scale separately and requires distinct representations for each scale. As expected, a significant performance improvement for late fusion over early fusion as shown in Table~\ref{tab:fusion_strategy}. In our early fusion approach, we include lateral connections across scales to encourage contextual information sharing. This performance gap validates our decision to perform per-scale temporal alignment, as temporal domain shifts at different frequencies are distinct. 

\subsubsection{$L2$ Loss.}
Scale coordination via $L2$ variance minimisation prevents any scale from dominating training. This ensures the fusion process does not overlook temporal information from any scale. Ablation of the $L2$ loss results in a 1.203\% performance loss, as seen in Table~\ref{tab:l2_ablataion}. This shows that scale coordination and the prevention of scale dominance improves the learned fusion representation. 

\subsubsection{Hyper-parameter Sensitivity.}
\label{sec: hyper-parameter_sensitivity}
We further examine the hyper-parameter sensitivity of our model to two key hyper-parameters: $L2$ loss weight and scale-spacing. In the federated setting, we are unable to access source and target data before hand, as such, the variance in the KL divergences cannot be pre-determined. As such, a model that is robust to a wider range of $L2$ loss weight would be adaptable to a wider range of domain shifts. Similarly, we cannot pre-determine the best scale-spacing as we are not privy to the full action-sequence length. A model robust to various scale spacings would allow for good performance even if a specific scale is not contributing as well as the other scales. As shown in Table~\ref{tab:hyper_sensitivity}, METAL's scale fusion approach is robust to various scale-spacing configurations, achieving similar performance. In the [1, 8, 64] configuration, we see that $S_3=64$, a very coarse scale, performs relatively poorly but the model is robust to this poorly performing scale. Different values of $L2$ loss weight is explored. While model performance is similar across the various values, our chosen value of 0.1 has the best performance. Dropping to 0.05 or increasing to 0.2 results in an identical performance plateau 52.67\%, while a heavy penalty of 0.5 finally begins to degrade the individual scale performances. 

\section{Conclusion}
In this work, we pioneer the exploration of the Federated Video Domain Adaptation (FVDA) task, which seeks to transfer knowledge from statistically heterogenous source-clients to the target-server while preserving data privacy in the video modality. We propose METAL, which learns multi-scale temporal representations and performs per-scale knowledge distillation to tackle the temporal domain shift. METAL preserves the distinct temporal representations by choosing a late fusion approach, allowing each scale to contribute unique temporal information to the fusion head through feature-based knowledge distillation. Scale coordination is further introduced to prevent the dominance of any scale during training. Extensive experimentation and detailed ablation and hyper-parameter studies across two challenging video domain adaptation datasets validate METAL's approach to FVDA. 



%
%
\bibliographystyle{splncs04}
\bibliography{main}
\end{document}